\documentclass[10pt,twocolumn,letterpaper]{article}

\usepackage{cvpr}
\usepackage{times}
\usepackage{epsfig}
\usepackage{graphicx}
\usepackage{amsmath}
\usepackage{amssymb}

\usepackage{xcolor}
\usepackage{algorithm2e,algpseudocode}
\usepackage{setspace}
\usepackage{amsfonts, amsmath, amssymb}
\usepackage{caption}
\usepackage{subcaption}
\usepackage{afterpage}

\usepackage[breaklinks=true,bookmarks=false]{hyperref}

\cvprfinalcopy 

\def\cvprPaperID{10436} 

\begin{document}

\title{ Progressive VAE Training on Highly Sparse and Imbalanced Data}

\author{Dmitry Utyamishev\\
University of Illinois at Chicago\\
1200 W Harrison St, Chicago, IL 60607\\
{\tt\small dutyam2@uic.edu}
\and
Inna Partin-Vaisband\\
University of Illinois at Chicago\\
1200 W Harrison St, Chicago, IL 60607\\
{\tt\small vaisband@uic.edu}
}

\maketitle

\begin{abstract}

   In this paper, we present a novel approach for training a Variational Autoencoder (VAE) on a highly imbalanced data set. The proposed training of a high-resolution VAE model begins with the training of a low-resolution core model, which can be successfully trained on imbalanced data set.  In subsequent training steps, new convolutional, upsampling, deconvolutional, and downsampling layers are iteratively attached to the model. In each iteration, the additional layers are trained based on the intermediate pretrained model -- a result of previous training iterations. Thus, the resolution of the model is progressively increased up to the required resolution level. In this paper, the progressive VAE training is exploited for learning a latent representation with imbalanced, highly sparse data sets and, consequently, generating routes in a constrained 2D space. Routing problems (e.g., vehicle routing problem, travelling salesman problem, and arc routing) are of special significance in many modern applications (e.g., route planning, network maintenance, developing high-performance nanoelectronic systems, and others) and typically associated with sparse imbalanced data. In this paper, the critical problem of routing billions of components in nanoelectronic devices is considered. The proposed approach exhibits a significant training speedup as compared with state-of-the-art existing VAE training methods, while generating expected image outputs from unseen input data. Furthermore, the final progressive VAE models exhibit much more precise output representation, than the Generative Adversarial Network (GAN) models trained with comparable training time. The proposed method is expected to be applicable to a wide range of applications, including but not limited image impainting, sentence interpolation, and semi-supervised learning. 
\end{abstract}

\section{Introduction}

\begin{figure}[t]
\begin{center}
   \includegraphics[width=0.9\linewidth]{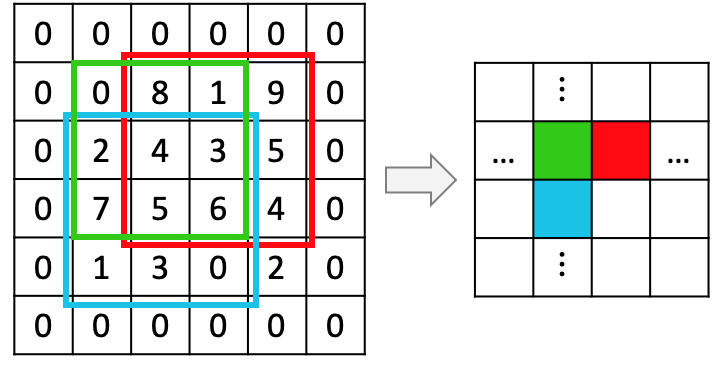}
\end{center}
   \caption{Illustration of local connectivity within convolutional layers – a CNN attribute that enables effective detection of local features in the input image.}
\label{fig:fig1}
\end{figure}

A Convolutional Neural Network (CNN) \cite{krizhevsky2012imagenet} is a machine learning (ML) architecture. Owing to local connectivity of convolutional layers, CNNs are commonly used for detecting complex local patterns within high-resolution 2D maps, as illustrated in Figure~\ref{fig:fig1}. In particular, image manipulation problems can be efficiently solved with CNNs \cite{isola2017image}. Training a CNN model is, however, a non-trivial problem.

\begin{figure}[t]
\begin{center}
   \includegraphics[width=\linewidth]{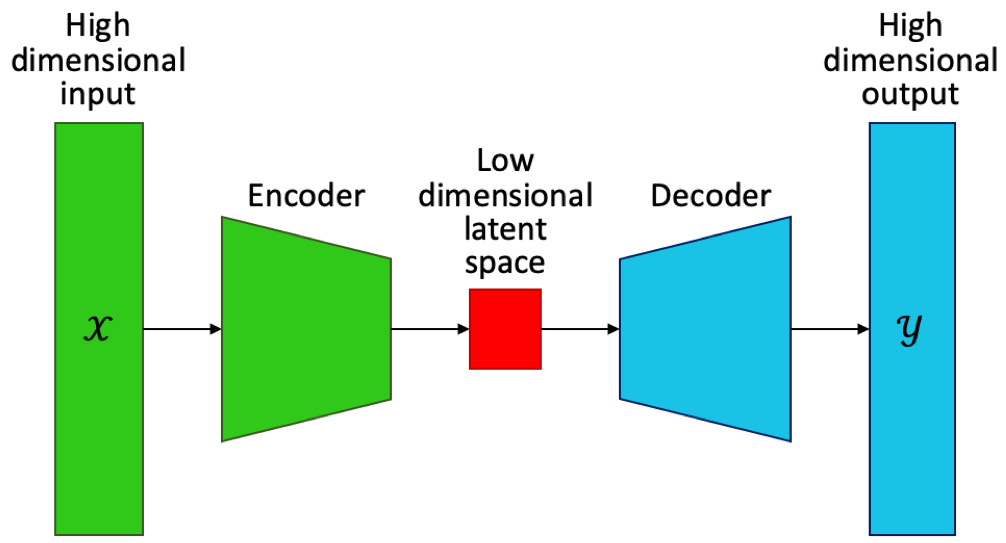}
\end{center}
   \caption{Stacked VAE.}
\label{fig:fig2}
\end{figure}

A typical CNN comprises multiple convolutional and deconvolutional layers separated with upsampling or downsampling layers. One of the most fundamental and common configurations of CNN is VAE – a CNN topology which enables the encoding of a high-dimensional 2D input into a low-dimensional inner representation. Consequently, the high-dimensional output image is reconstructed solely based on the low-resolution inner representation. The structure of a stacked autoencoder is illustrated in Figure~\ref{fig:fig2}.

In its simplest form, the multidimensional VAE input and output images are identical. In this case, VAE acts as an image compression tool. Alternatively, more complex VAE configurations make processing of the input image possible by mapping one image onto another. For advanced image-to-image transformations, a reasonably deep neural network is required, yielding image processing with complex non-linear logic. Increasing the number of VAE layers, however, poses new challenges on the training process. For example, the vanishing gradient problem is a primary concern in deep VAE networks. This problem is a direct consequence of backpropagation (\ie, gradient descent) algorithm, which minimizes the error function, 
$J(\Theta): \mathbb{R}^n \rightarrow \mathbb{R}$,
by iteratively updating the network model weights, 
$\Theta \in \mathbb{R}^n$,
in opposite direction to the gradient of the error function with respect to the network weights, 
$\Theta := \Theta - \eta \Delta J $. 
Here $\eta$ is the learning rate of the algorithm. Intuitively, the gradients become smaller with the increasing model accuracy. These small gradients tend to further decrease through continuous matrix multiplications in those inner network layers, significantly impeding the model training \cite{hochreiter1998vanishing}. 

\begin{figure}[t]
\begin{center}
   \includegraphics[width=0.9\linewidth]{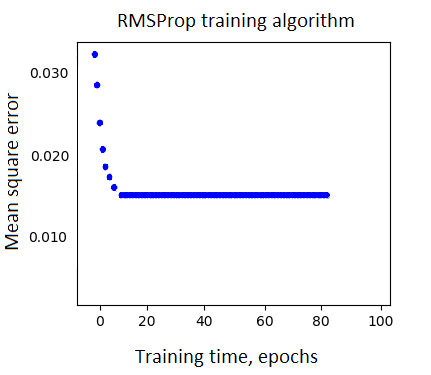}
\end{center}
   \caption{Illustration of model training,  saturating at a local minimum.}
\label{fig:fig3}
\end{figure}

Furthermore, owing to the high dimensionality of the data and, accordingly, the large number of learning parameters, those early layers of a neural network (as shown by the green shaded substructure in Figure \ref{fig:fig2}) are trained significantly slower than the later layers (as shown by the blue shaded substructure in Figure \ref{fig:fig2}), intensifying the deep training problem. In particular, the surface of the error function becomes flatter with increasing number of weights \cite{ruder2016overview} and vanishing gradients, further decreasing the training speed \cite{hochreiter1998vanishing}. Finally, backpropagation with a flat error function often saturates in a local minimum, as it is illustrated in the Figure~\ref{fig:fig3}. Training speed and convergence with complex deep VAE architectures is, therefore, a primary concern and the main focus of this paper.

\section{Related work}

Several methods have been proposed to mitigate the issue of training speed and convergence. 
For example, reasonable training speedup and convergence has recently been demonstrated with ReLU-based activation functions in deep neural networks. The traditional sigmoid activation function saturates quickly at both, very positive and negative argument values, yielding vanishing gradient values within these argument ranges. Alternatively, with the ReLU activation function, $f(x) = \max(0,x)$, first derivative is unity when function argument is positive. Thus, ReLU \cite{agostinelli2014learning} activation function exhibits non-linear behavior (as required for complex transformations in neural networks) and no signal degradation through multiple neural network layers. Alternatively, the zero derivative of ReLU function for a negative argument increases the gradient vanishing probability for the negative inputs.  Saturation of backpropagation at zero local minimum is, therefore, still a primary concern with ReLU activation function.

Momentum-based training is yet another approach to speed up deep training and mitigate the saturation in a local minimum with sparse input data. The effectiveness of the traditional gradient descent algorithms degrades with increasing sparsity of the training data \cite{ruder2016overview}. Ideally, suboptimal local minima should be avoided. In practice, the error function is, however, often a complex non-convex surface with saddle points surrounded by flat regions (\ie, constant error function). Escaping these regions is a significant challenge for the gradient descent algorithms.  Momentum-based functions, such as Adadelta \cite{zeiler2012adadelta} and RMSprop \cite{ruder2016overview} have been demonstrated to accelerate gradient descent convergence and mitigate oscillations of the algorithm caused by ravines slopes around local minima. With these techniques, momentum is accumulated for those parameters with similar gradient direction. As a result, oscillation is reduced, and network model converges faster. 

Sophisticated methods with adaptive learning rates have also been considered. With these methods, deep learning training converges faster with default network parameters, eliminating the need for manually adjusting the learning rates. As a result, deep training performance with sparse input data is significantly increased with these approaches.

Yet another approach commonly used with computer vision problems is transfer learning \cite{torrey2010transfer}. To mitigate the complexity, transfer learning approaches heavily rely on pretrained reference models. A common practice is to lock majority of the pretrained model layers associated with fundamental low-level features, attach new layers associated with an application/object-specific specialized feature, and retrain, repurposing the learning features of the reference model. To effectively leverage transfer learning with sparse data sets, the number of locked layers should be increased with the increasing sparsity of data. At the limit, all the pretrained model layers are locked and only the additional new layers are trained. 

The method proposed in this paper borrows from the transfer learning approach, as described in Section~\ref{Method}. The proposed method is experimentally verified and compared with existing state-of-the-art methods, yielding a significant increase in training speedup and performance, as described in Section~\ref{Experimental data}.

\section{Method}\label{Method}
\subsection{ML objective}
A general image impainting problem is formulated in this section as a supervised ML task. Let $\mathcal{X}$ and $\mathcal{Y}$ be the set of, respectively, learning objectives and the corresponding impainting solutions. For a single ML objective, $x \in \mathcal{X}$, the corresponding output image, $y_x \in \mathcal{Y}$, is an $n \times n$ bitmap of pixels indexed, $t_{i,j}, 1 \leq i, j \leq n$. Each pixel within an output image, $y_x$ is associated with a binary score, $y_{x_{i,j}} = 0$ or $y_{x_{i,j}} = 1$ if the pixel $t_{i,j}$ is, respectively, excluded from or included within the output image.

The primary goal is to train a ML system $f(\mathcal{X}, \Theta)$ that provides the conditional probability of each pixel, $t_{i,j}$, to be either included within (\ie, $f_{i,j} \geq 0.5$) or excluded from (\ie, $f_{i,j} < 0.5$) the preferred ML solution, 
\begin{equation}
    f_{i,j} (\mathcal{X}, \Theta) = P (t_{i,j} = 1 \vert \mathcal{X}), 
\end{equation}
where $\Theta$ is the trained model of ML weights. The training data set $\{(\mathcal{X}_i,\mathcal{Y}_i )\}_{i=1}^N$ comprises $N$ synthetic ML objectives in the bitmap representation and $N$ corresponding reference image outputs (\ie, the true labels). Note that the model $\Theta$ is trained based on the joint probability distribution of the input features, $\mathcal{X}_i$, and output observations, $\mathcal{Y}_i$, yielding a generative network. 

\subsection{Concept overview}
In many image-to-image transformations, training sets are available with various resolution levels. While models trained on high-resolution data often suffer from gradient vanishing and other limitations (see Section 1) those lower-resolution models typically exhibit high performance and convergence due to lower number of layers and model parameters \cite{genevay2017gan,hochreiter1998vanishing,ruder2016overview}.
The proposed solution leverages iterative training with gradually increasing dimensionality as well as the principles of repurposing a pretrained model, as demonstrated with transfer learning methods. 

\begin{figure}[t!]
\begin{center}
   \includegraphics[width=\linewidth]{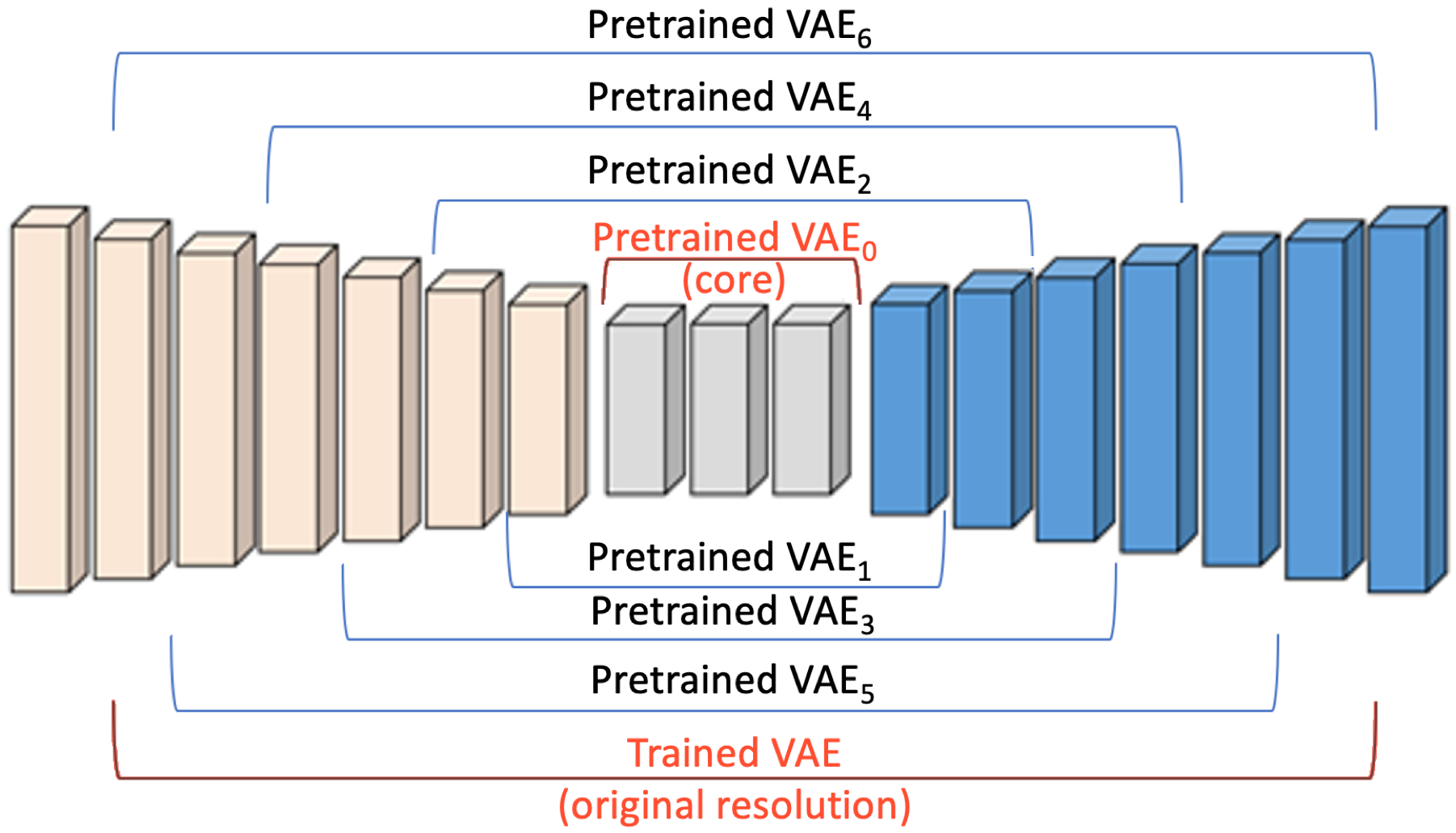}
\end{center}
   \vspace{-5pt}
   \caption{Illustration of progressive VAE training. The desired, high-resolution VAE shares layers with the intermediate autoencoders and the most inner VAE (as shown with grey shade).  } \vspace{-10pt}
\label{fig:fig4}
\end{figure}

To train a stacked deep VAE, an inner low-dimensional substructure of the autoencoder is first identified (referred here as core VAE). The objective in the first iteration is to determine an inner substructure, which (due to lower dimensionality of the inner VAE layers) enables successful model training on sparse data. Once the core autoencoder is identified, the outer, high dimensional layers are stripped, and the core model is successfully trained on a low dimensional data set. In consequent iterations, the stripped layers are progressively added back, and the model is retrained on training sets with increasingly higher resolutions. Finally, a complete VAE is trained on the original, high-resolution data set. This approach is illustrated in Figure~\ref{fig:fig4} with a eight-step training process, yielding the lowest resolution core network, $\text{VAE}_0$ (as shown with grey shade), intermediate networks, $\text{VAE}_1 \ldots \text{VAE}_6$ with progressively higher resolutions, and the final high-resolution network $\text{VAE}$, comprising layers shared among all the networks, $\text{VAE}_0 \ldots \text{VAE}_6$.

\subsubsection{VAE loss function}
Mean square error (MSE) loss function is typically used with autoencoders for evaluating sum of squared distances between the predicted values and true labels \cite{rao1980some}. This loss function is, however, less effective with highly imbalance data, where an empty output exhibits a small MSE, thus yielding a legitimate solution. In a typical routing problem, an expected routing output with a single, often quite short path, exhibits high similarity with an empty (\ie, no path) solution, yielding low routability with MSE metric. The problem escalates with increasing input resolution, further increasing the VAE training complexity and degrading the accuracy of the trained models.

To account for specifics of routing problems with an unbalanced data set, a custom loss function is proposed. Note that this function is relevant with other ML problems, such as image impainting, super resolution, and style transfer \cite{ledig2017photo,zhao2016loss,mathieu2015deep,deshpande2017learning,wang2016non}. The proposed loss function is designed to penalize the model if the number of tiles, $n_t$, included by the model within a routing path is different from the number of tiles in a reference routing path, $n_{t,ref}$. The penalties for $n_t$ exceeding and falling short of $n_{t,ref}$ differ. A path with redundant tiles is not optimal in terms of its length. It also reduces the routing capacity of the 2D space beyond the expected. Yet, such a path is considered to be legal if it connects all the input/output pins. Alternatively, if $n_t < n_{t,ref}$ and the reference path is optimal, some components in the model solution are disconnected and the path is, therefore, incorrect. In particular, the $n_t < n_{t,ref}$ penalization pertain to the “all-zeros” local minimum. The proposed loss function accounts for $\vert n_t - n_{t,ref} \vert \ne 0$ with penalty rate of $k_\text{sub-opt}$, and for incomplete routes with additional penalty rate of $k_\text{err}$, yielding the following loss function for a predicted routing output, $\hat{y}$, and a reference routing solution, $y^\text{ref}$,
\begin{equation}
    \label{eq:f_loss}
    f_\text{loss} = \text{MSE}\left(\hat{y}, y^\text{ref} \right) \cdot \left( 1 + k_\text{sub-opt} \cdot \text{step} \cdot \text{distance} \right),
\end{equation}
where\\
\begin{equation}
    \label{eq:distance}
    \text{distance} = \sum_{i,j} H(\hat{y}_{i,j}) - \sum_{i,j} H(y_{i,j}^{ref}) 
\end{equation}
\begin{equation}
    \label{eq:step}
    \text{step} = k_\text{err} \cdot  \text{sign}(\text{distance} - 1)  + 1.
\end{equation}
Here $H()$ is the Heaviside \cite{weisstein2002heaviside} step function. In this paper, the proposed loss function is used with $k_\text{sub-opt} = 10^{-3}$ and $k_\text{err} = 10^2$.

\subsubsection{Formal training approach }

The progressive training process starts with identifying the core autoencoder $\text{VAE}_0$. This core VAE is trained with a training set of resolution $2n \times 2n$. After training, the layers of $\text{VAE}_0$ are locked and $\text{VAE}_1$ is trained with training set of resolution $2^2n \times 2^2n$. Note, that the inner layers of $\text{VAE}_1$ are already trained and locked and only the outer layers are trained. After all intermediate $\text{VAE}_i$ layers are trained with the individual training sets of resolution $2^{1+i}n \times 2^{1+i}n$, the final VAE is trained with the desired resolution. The pseudocode of the training algorithm is described in Algorithm \ref{algo:flow}.
 
\begin{algorithm}
  \onehalfspacing
  \SetKwFunction{Construct}{Construct}
  \SetKwFunction{Train}{Train}
  VAE $v_0$ = train the core autoencoder\;
  list $models$ = \Call{Construct}{$v_0$}\;
  \Call{Train}{$models$}\;
  \singlespacing
  list  
  \Construct{\textnormal{VAE} $v_0$}{\\
    \While{\textnormal{!(the required model resolution)}}{
        \nl Detach the first convolutional and down sampling 
        layers and the last upsampling and deconvolutional layers. Consider the remaining structure throughout the algorithm.\\\vspace{5pt}
        \nl Attach a pair of new convolutional and corresponding downsampling layers (in this order) to the current most outer input layer. \label{enum:2} \\\vspace{5pt}
        \nl Attach a pair of new upsampling and deconvolutional layers (in this order) to the current most outer output later. \label{enum:3}\\\vspace{5pt}
        \nl Attach a new three-channel input and one-channel output convolutional layers to layers described in steps \ref{enum:2} and \ref{enum:3}. \label{enum:4}\\\vspace{5pt}
        \nl Append the outermost layers described in step \ref{enum:4} to the $models$ list. \\
    }
}
\vspace{10pt}void 
\Train{\textnormal{list} $models$}{\\
    \For{\textnormal{each $model$ in $models$}}{
        \If{\textnormal{training data for $model$ is available}}{
           Train $model$ with the training data\;
        }
    }
}
\vspace{15pt}
\caption{Progressive construction of VAE substructures.}
\label{algo:flow}
\end{algorithm}

Preliminary route-free training is another technique developed in this work for mitigating the probability of convergence to the “all-zeros” local minimum. For this purpose, a route-free training set is generated. The true label for each data point in this set comprises the tiles indices of the input pins (and no additional routed tiles). Intuitively, this route-free training promotes inclusion of the pin tiles within the routing path. As a result, the solution space is shifted towards the non-flat regions of the loss function hyperplane, reducing the probability of the convergence to “all-zeros” solution. 

\begin{figure*}[t]
    \centering
    \begin{subfigure}[]{0.31\textwidth}
        \includegraphics[width=\columnwidth]{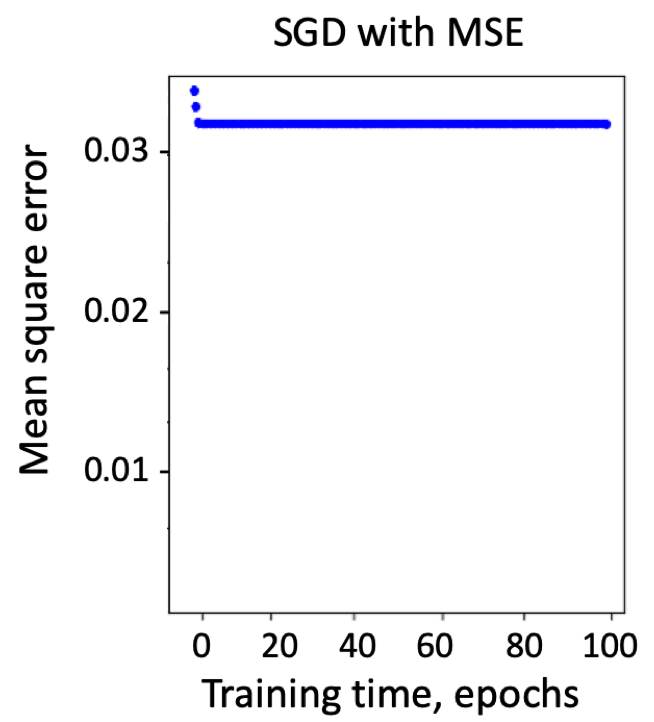}
        \caption{}
        \label{fig:training_conv_a}
    \end{subfigure}
    \begin{subfigure}[]{0.31\textwidth}
        \includegraphics[width=\columnwidth]{Figures/training_conv_a}
        \caption{}
        \label{fig:training_conv_b}
    \end{subfigure}
    \begin{subfigure}[]{0.33\textwidth}
        \includegraphics[width=\columnwidth]{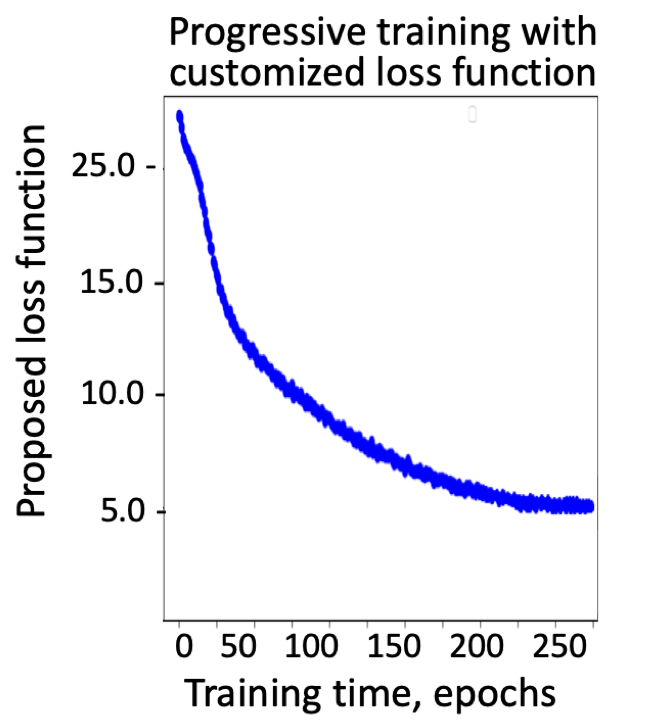}
        \caption{}
        \label{fig:training_conv_c}
    \end{subfigure}
    \caption{Model training with various training methods and loss functions. (a) SGD optimization with MSE loss function, saturating at a local minimum. (b) RMSProp optimization (with learning rate of $10^{-4}$) with MSE, saturating at a local minimum. (c) Training of the outermost layers of the desired high-resolution model with progressive VAE, custom loss function, $f_\text{loss}$, (see (\ref{eq:f_loss})(\ref{eq:distance})(\ref{eq:step})), and preliminary training, converging to correct routing solution.}
    \label{fig:training_conv}
\end{figure*}

Model training with stochastic gradient descent (SGD) and RMSProp is shown in Figure \ref{fig:training_conv} with the MSE function and the customized $f_\text{loss}$ function (see (\ref{eq:f_loss})(\ref{eq:distance})(\ref{eq:step})), exhibiting the convergence trends, as described in this section.

\section{Experimental data}\label{Experimental data}
\subsection{Experimental results}
\begin{figure}[t]
    \centering
    \begin{subfigure}[]{\columnwidth}
        \includegraphics[width=\columnwidth]{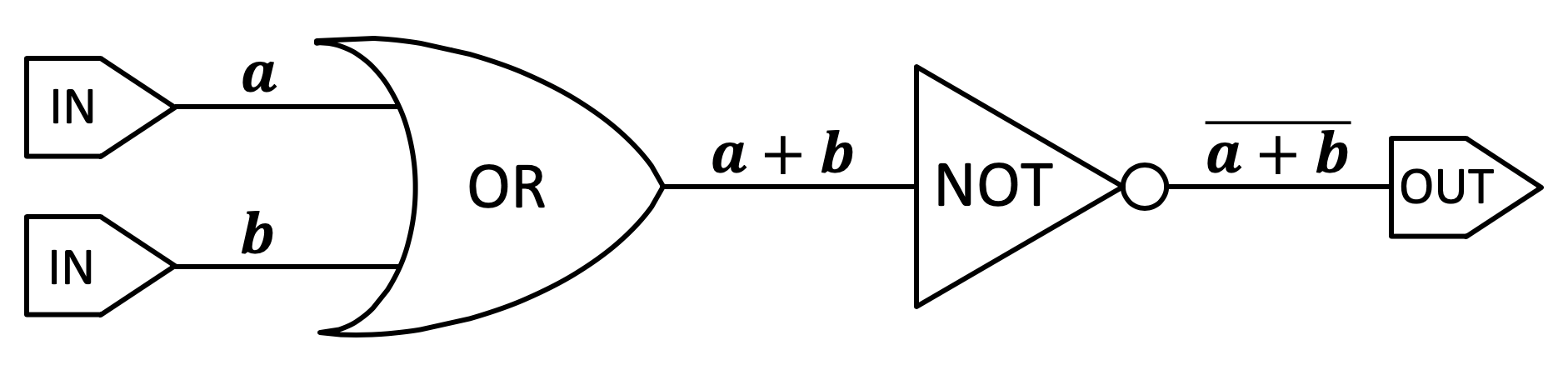}
        \caption{}\vspace{10pt}
        \label{fig:schem_a}
    \end{subfigure}
    \begin{subfigure}[]{\columnwidth}
        \includegraphics[width=\columnwidth]{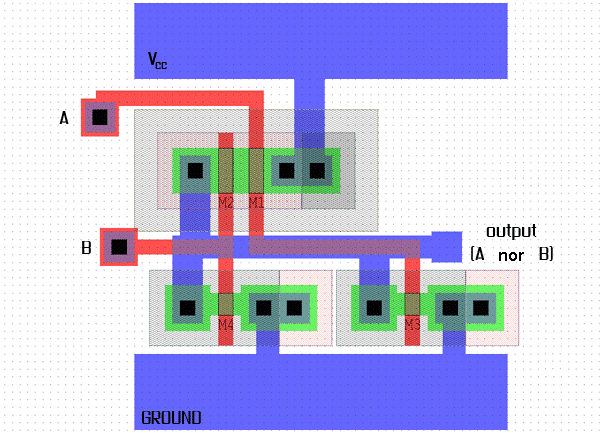}
        \caption{}
        \label{fig:schem_b}
    \end{subfigure}
    \caption{Schematic of a basic Boolean function $NOR = \overline{(a+b)}$, (a) logic representation, where the connections between the gates are the logic nets of the system, and (b) physical implementation in an electronic device. The physical wires connecting functional circuits are shown in red, exhibiting a non-trivial path for this simple two-pin component.  }
        \label{fig:schem}
\end{figure}

The proposed approach is demonstrated on a practical routing problem \cite{kahng2011vlsi}. Routing is a major phase of electronic circuit design process. During this phase, the electronic components that have previously been placed within restricted space, are connected with physical wires with respect to their intended functionality. For example, to implement a Boolean function $NOR = \overline{(a+b)}$ (inversion of ‘$a$’ OR ‘$b$’), signal pins ‘$a$’ and ‘$b$’ are connected with logic nets to the individual inputs of gate ‘$OR$’, the output of the gate ‘$OR$’ is connected with a net to the input of gate ‘$NOT$’, and the output of the gate ‘$NOT$’ is connected to the output pin of the system, as illustrated in Figure~\ref{fig:schem}. 

During the routing process, all the logic nets are implemented as physical routes within the technology constraints (\eg, wires can only be routed in vertical and horizontal directions, but not horizontally). The wire route within a constrained area with limited net capacity exhibits a complex path even with this simple two-pin function. Alternatively, modern microprocessors comprise billions of Boolean gates and complex technology and routing constraints \cite{moffitt2008coming}. Routing in these systems is a NP-hard problem and a critical challenge for next generation high-performance nanoelectronic systems \cite{du2013advances}.

In this paper, an input image is represented by array of pixels. Each tile exhibits several characteristics, such as color channels and other special constraints. The input of a image impainting problem is a set of per-pixel nts and a set of all the logic nets, as defined by the physical input and output pin positions. The output of the routing problem is a set of tiles that should be included within the routing paths of the individual input nets. Note, that optimal routing of a net with more than two pins (\eg, a net that connects the output of an inverter to inputs of two other inverters) is a NP-hard problem.
Traditionally routing problems are solved with approximation methods, yielding suboptimal solution \cite{dai2011nctu}. We propose to solve this critical problem by formulating it as a ML problem. Consider the following definitions. 

A general routing problem is mapped here onto the classical image impainting problem, as defined in Section 3. Let $\mathcal{X}$ and $\mathcal{Y}$ be the set of, respectively, single-path routing objectives (as defined by all the starting and ending points of a single path) and the corresponding single-net routing solutions. A routing objective is to find the preferred routing path of tiles, $\mathcal{Y}_x \in \mathcal{Y}$, connecting a certain number of placed pins within a given vertical and horizontal per-tile capacity, as defined by $x \in \mathcal{X}$. For any routing objective, $x \in \mathcal{X}$, the corresponding single-net routed output image, $y_x \in \mathcal{Y}$, is an $n \times n$ bitmap of tiles indexed with their physical locations, $t_{i,j}, 1 \leq i, j \leq n$. Each tile within an output image, $y_x$, is associated with a binary score, $y_{x_{i,j}} = 0$ or $y_{x_{i,j}} = 1$ if the routing tile $t_{i,j}$ is, respectively, excluded from or included within a routing path.

\begin{figure*}[t]
    \centering
    \includegraphics[width=0.9\textwidth]{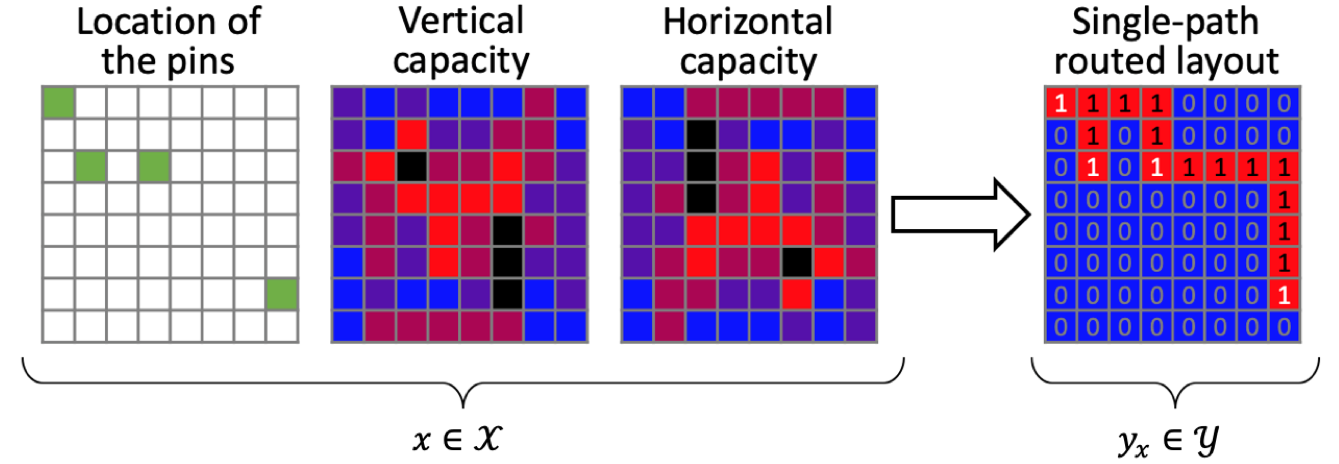}
    \vspace{15pt}
    \caption{Illustration of a single net routing objective $x \in \mathcal{X}$ and the corresponding single-net routed output $y_x \in \mathcal{Y}$.  }
    \label{fig:in_and_out_fake}
\end{figure*}

\begin{figure*}[t]
    \centering
    \begin{subfigure}[]{0.3\textwidth}
        \includegraphics[width=\columnwidth]{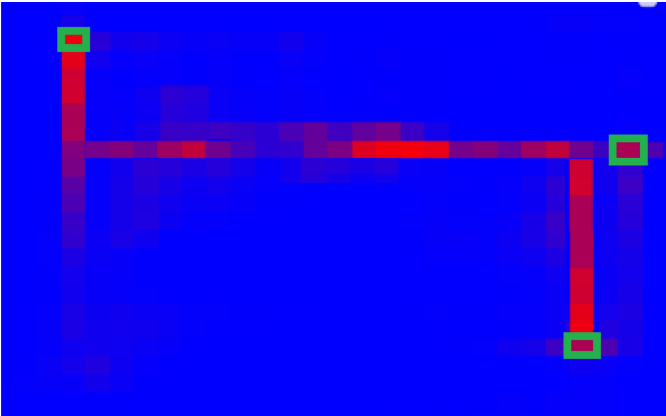}
        \caption{}
        \label{fig:(in)correct_a}
    \end{subfigure}
    \begin{subfigure}[]{0.3\textwidth}
        \includegraphics[width=\columnwidth]{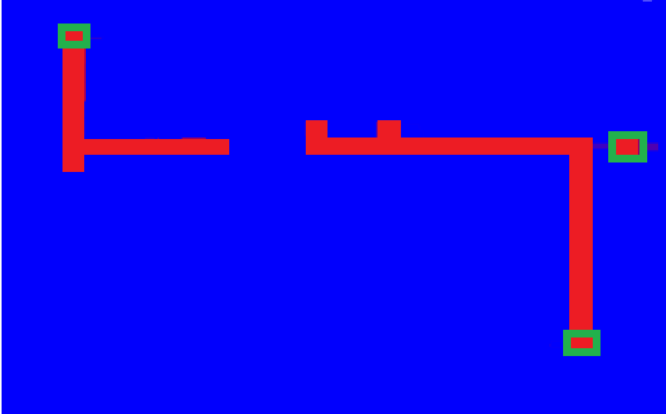}
        \caption{}
        \label{fig:(in)correct_b}
    \end{subfigure}
    \begin{subfigure}[]{0.32\textwidth}
        \includegraphics[width=\columnwidth]{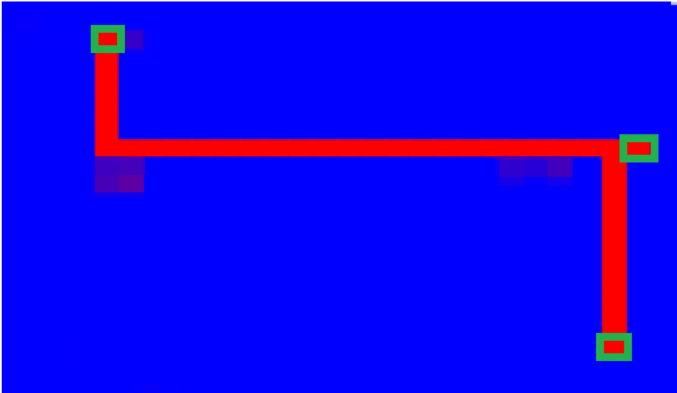}
        \caption{}
        \label{fig:(in)correct_c}
    \end{subfigure}
    \caption{A typical output of ML based router, (a) a blurry path is generated with a conventional neural network, (b) the output of (a) is refined with the thresholding function, yielding a disconnected and, thus, illegal path,  and (c) a legal routing path.}
    \label{fig:(in)correct}
\end{figure*}

Note that a primary objective is maximizing the routability (\ie, the number of routed paths), while minimizing the overall wirelength (\ie, the total number of tiles included within all the routed paths) of the routing solution. 
For each routed wire, the number of tiles included within a path, $O(n)$, is significantly lower than the total number of tiles, $n^2$, yielding a highly sparse and imbalanced data. These definitions are illustrated in Figure~\ref{fig:in_and_out_fake}.
An example of correct and incorrect routing solution is shown in Figure \ref{fig:(in)correct}.

Synthetically-generated exhaustively-routed training set was produced with an auxiliary state-of-the-art router \cite{pan2012fastroute} and a regular VAE model was trained with RMSprop \cite{ruder2016overview} training method. With this approach, training process has not successfully converged, but saturated at a local minimum. The output of this model is, therefore, of limited use to attain routing of high performance computing devices. Due to sparsity of the data, the most common outcome of the training is the saturation of the model in a local minimum that corresponds to an empty output with all output tiles in the same class, $y_{i,j} = 0, \forall i,j$. 
Routing problem is selected in this paper as a demonstration vehicle due to its high impact on the microprocessor industry and our ability to perform perceptual evaluation of the results in addition to synthetic ML performance metrics. To verify the correctness of the models, individual routing solutions are traversed with BFS algorithm, evaluating the connectivity of all the pins and nets. Number of successfully routed nets is used as a main metric due to its qualitative (\ie, indicates the system routing ability) and quantitative characteristics.
If model tends to include less amount of tiles, that is needed, pins may be not connected, and number of the routed net will be lower, than reference. Similarly, if model tends to include more tiles in the routing path, capacity of design saturates faster, that leads to unavailability to route later nets. 
In this paper, the routability of the individual models is measure as per cent of routability of a state-of-the-art deterministic router, FastRoute 4.1.

\subsection{Implementation and performance comparison}

ISPD’98 $64 \times 64$ resolution benchmark ‘ibm02.modified’ \cite{alpert1998ispd98} is used for evaluation. With three features per tile (vertical and horizontal available net capacity and binary pin metric), a total of $64 \times 64 \times 3$ features are considered, yielding a $64 \times 64 \times 3$ dimensional input. Similarly, a $64 \times 64 \times 1$ dimensional output space is required to describe the inclusion or exclusion of each tile from the individual routing solutions. Architecture and detailed description of each layer of the proposed VAE is illustrated in the Figure~\ref{fig:arch}.
A simple VAE, GAN, and progressive VAE network are implemented and evaluated with this benchmark. All networks described in this section are prototyped in Python 3.7 using Keras neural-network library \cite{gulli2017deep} with Tensorflow backend.

\subsection{Variational autoencoder}

\begin{figure*}[t]
    \centering
    \begin{subfigure}[]{0.5\columnwidth}
        \includegraphics[width=\columnwidth]{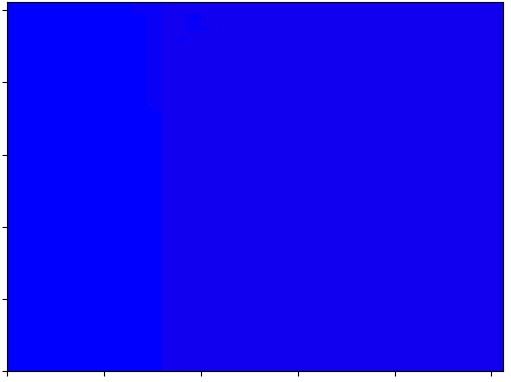}
        \caption{}
        \label{fig:route_out_a}
    \end{subfigure}
    \begin{subfigure}[]{0.5\columnwidth}
        \includegraphics[width=\columnwidth]{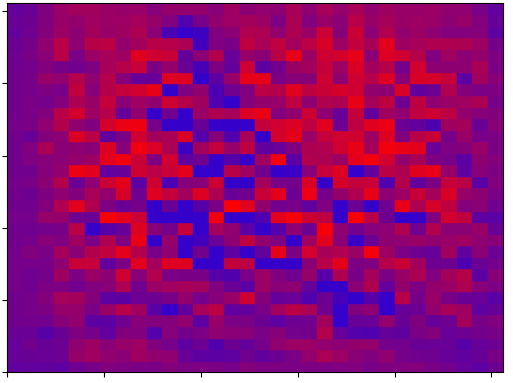}
        \caption{}
        \label{fig:route_out_b}
    \end{subfigure}
    \begin{subfigure}[]{0.5\columnwidth}
        \includegraphics[width=\columnwidth]{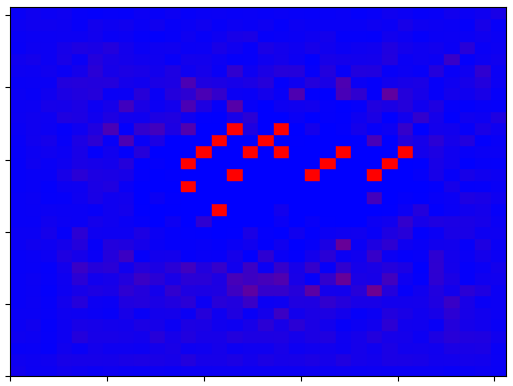}
        \caption{}
        \label{fig:route_out_c}
    \end{subfigure}
    \begin{subfigure}[]{0.49\columnwidth}
        \includegraphics[width=\columnwidth]{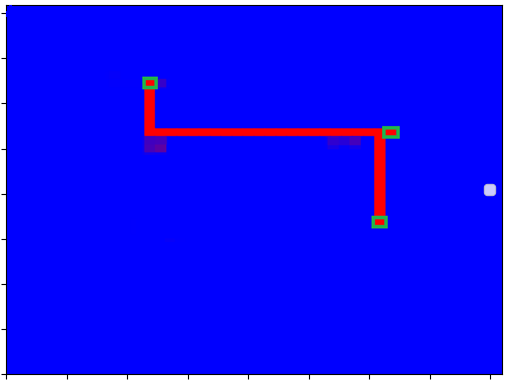}
        \caption{}
        \label{fig:route_out_d}
    \end{subfigure}
    
    \caption{Routing output of various neural networks trained on the same synthetic data set and evaluated on a typical unseen three-pin net, (a) simple VAE, (b,c) GAN in two stable states, and (d) progressive VAE.}
    \vspace{5pt}
    \label{fig:route_out}
\end{figure*}

Stacked VAE is designed to solve the critical problem of routing in nanoelectronic devices by utilizing ML imaging methods and parallelization provided by GPU platforms. Input and output spaces are defined as described in Section 3.  
This model is trained on the training set of 12,000 nets defined within the $64 \times 64$ routing space. Input nets are generated with random pin and obstacles locations. Here synthetic obstacles are used to analogize the already placed wires, fully or partially occupying the available net capacitance at certain tiles. The reference outputs (\ie, truth labels) are obtained with an auxiliary state-of-the-art router FastRoute 4.1. The proposed VAE is trained on this synthetic model. Fastest convergence rate is observed with the RMSprop method with cyclical learning rate. The trained VAE is however unable to predict correct routing solutions neither on new, previously unseen nets, nor on those nets selected from the training set. In all these cases, a trivial output of empty wiring paths is produced by the trained autoencoder. Example of a typical routing output with this method is shown in Figure~\ref{fig:route_out_a}. Benchmark routability with simple VAE is, therefore, zero per cent of the routability with the reference router on the same benchmark \cite{alpert1998ispd98}. Note those trivial cases where all pins are placed in the same tile and no routing is required are excluded from the routability calculations.
 
\subsection{Generative Adversarial Network}
\begin{figure*}[t!]
    \begin{center}
      \begin{subfigure}[]{\textwidth}
        \includegraphics[width=\textwidth]{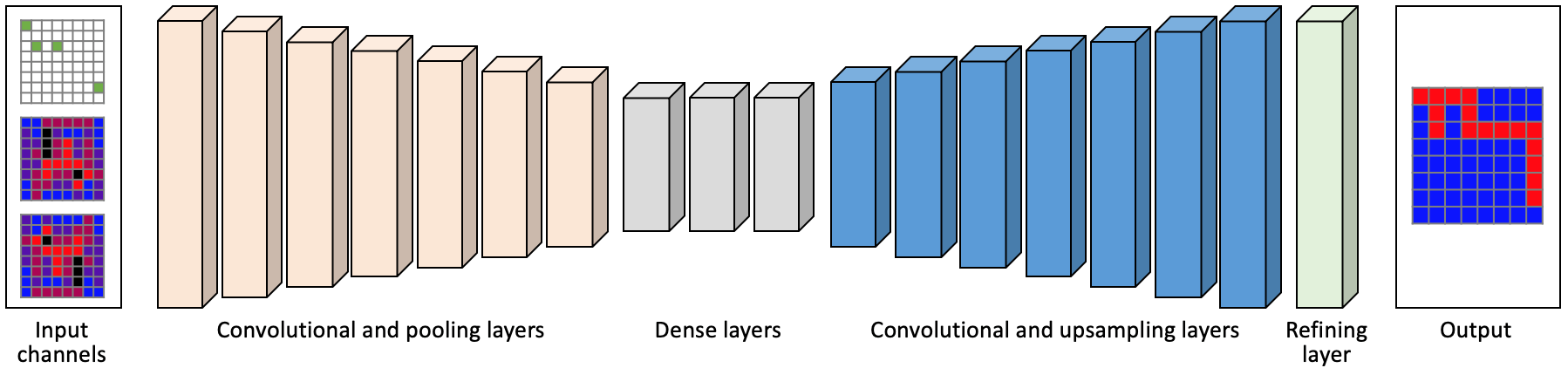}
        \caption{}\vspace{15pt}
        \label{fig:figarcha}
      \end{subfigure}
      \begin{subfigure}[]{\textwidth}
      \centering
        \includegraphics[width=0.90\textwidth]{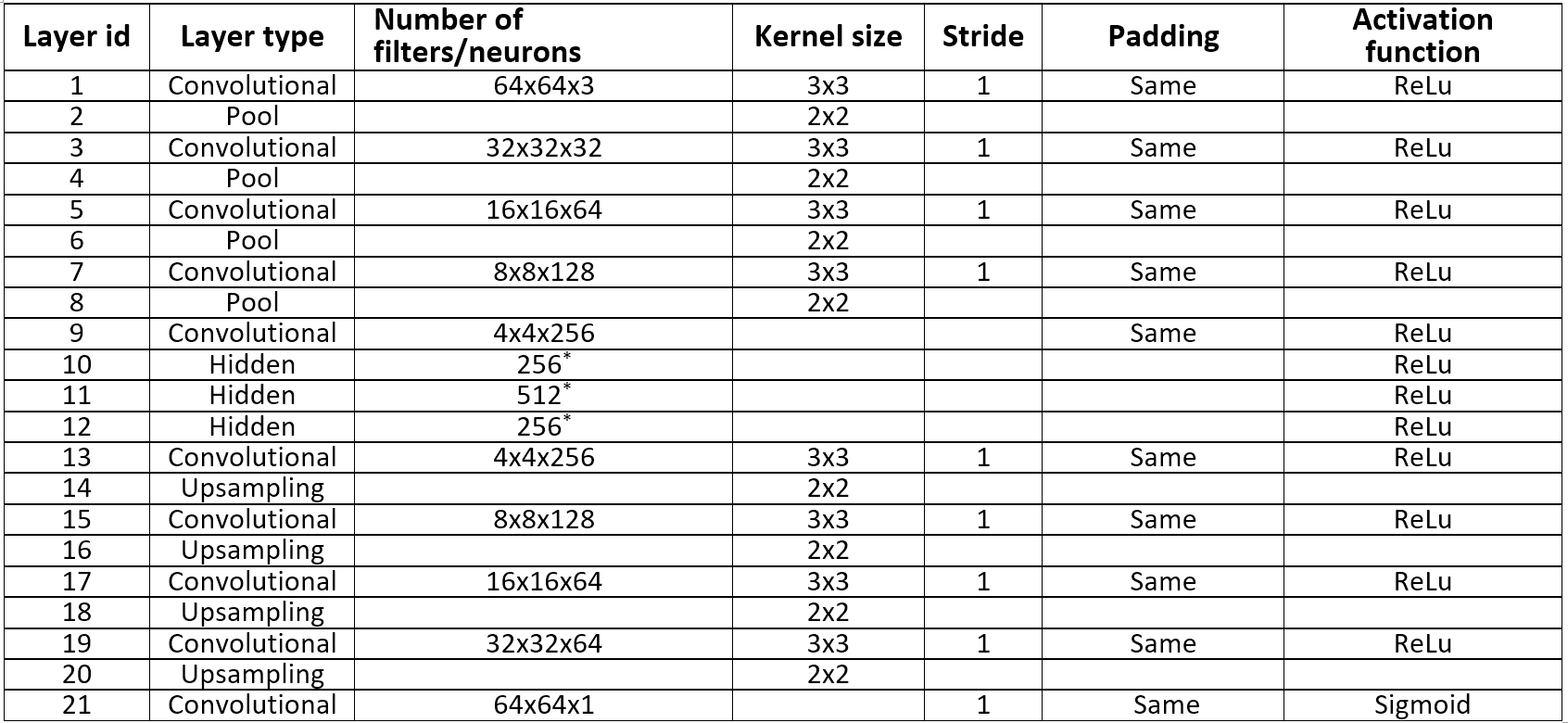}
        \caption{}
        \label{fig:figarchb}
      \end{subfigure}
    \end{center}
    \small *25\% dropout \cite{srivastava2014dropout} (a fraction rate of input units set to zero at each update during the training to prevent overfitting)  
   \caption{Architecture of the proposed generative network, (a) block level schematics comprising convolutional, dense, and deconvolutional layers, as well as an example of input features and impainting output, and (b) neural network parameters for each layer of the $64 \times 64$ router.}
\label{fig:arch}
\end{figure*}

GAN systems are commonly used to generate previously unseen images. With these methods, training is approached as a minimax game with a generating and discriminating mechanisms. The objective of the generator is to produce output similar to the reference . Alternatively, the objective of the discriminator is to distinguish between generated outputs and reference outputs. The structure of the proposed GAN is shown in Figure \ref{fig:arch}.
The network is trained on the same input data as the simple autoencoder. The training process has not converged after 10,000 iterations, yielding a total runtime of ten hours on Nvidia GTX1080 GPU. Oscillations between two local minima as observed \cite{goodfellow2016nips}, exhibiting a typical mode collapse behavior of a GAN system \cite{srivastava2017veegan}. As a result only few modes of a multimodal data are generated, producing perceptually variable routing outputs, as illustrated in Figures~\ref{fig:route_out_b} and ~\ref{fig:route_out_c}. All the results, however, exhibit low performance based on the routability metric and perceptual evaluation. A total of 2.7\% of all non-trivial nets are routed with this network, showing better result, than VAE router (2.7\% non-trivial net routed), but still not enough for using it in real applications.

\subsection{Progressive VAE}
The proposed progressive VAE network is designed to utilize all available high, intermediate, and low-resolution training sets. Note that generation of reference net routes for model training exhibits high time complexity with data resolution. For example, synthesizing exhaustively-routed, $64 \times 64$-resolution training set is significantly more time consuming than synthesizing 4 training sets with resolutions of $64 \times 64$, $32 \times 32$, $16 \times 16$, and $8 \times 8$. The low-resolution core router comprises same layers, as layers 7-15 in Figure~\ref{fig:arch} and an additional $8 \times 8 \times 1$ convolution output layer. This router is trained on an $8 \times 8$ data set. With every progressive VAE iteration, an additional convolutional and pool layer is attached to the input of the pretpre-rained VAE network and an additional upsamping and deconvolutional layer is appended at the end of the network. Finally, an output $i \times i \times 1$ layer is added to convert the dimensionality of the last convolutional layer, $i \times i$, to the routing output dimensionality of one. 

\section{Conclusion}
This research introduces a new approach for iteratively training VAE on highly sparse imbalanced data with progressively increasing training data resolution. The proposed method has been evaluated on routing benchmarks \cite{alpert1998ispd98}, successfully generating routes between placed pins in a constrained 2D space with limited routing capacity. The proposed method exhibits faster convergence and 96\% routability, as compared with 0\% and 2.7\% routability with simple VAE and GAN networks. 

\newpage

{\small
\bibliographystyle{ieee_fullname}
\bibliography{refs}
}

\end{document}